\documentclass{bmvc2k}

\usepackage{amsmath}
\usepackage{hyperref}       
\usepackage{color, soul}

\title{High Frequency Residual Learning for Multi-Scale Image Classification}

\addauthor{Bowen Cheng}{bcheng9@illinois.edu}{1}
\addauthor{Rong Xiao}{rongxiao@gmail.com}{2}
\addauthor{Jianfeng Wang}{jianfw@microsoft.com}{3}
\addauthor{Thomas Huang}{t-huang1@illinois.edu}{1}
\addauthor{Lei Zhang}{leizhang@microsoft.com}{3}

\addinstitution{
 University of Illinois at Urbana-Champaign\\
 Urbana, IL, USA
}
\addinstitution{
 Ping An Property\&Casualty Insurance Company of China\\
 China}
\addinstitution{
 Microsoft\\
 Redmond, WA, USA
}

\runninghead{Cheng et al}{High Frequency Residual Learning}

\newcommand{\eg}{\emph{e.g. }}

\newcommand{\ve}[1]{\mathbf{#1}} 

\begin{document}

\maketitle

\begin{abstract}
We present a novel high frequency residual learning framework, which leads to a highly efficient multi-scale network (MSNet) architecture for mobile and embedded vision problems. The architecture utilizes two networks: a low resolution network to efficiently approximate low frequency components and a high resolution network to learn high frequency residuals by reusing the upsampled low resolution features. With a classifier calibration module, MSNet can dynamically allocate computation resources during inference to achieve a better speed and accuracy trade-off. We evaluate our methods on the challenging ImageNet-1k dataset and observe consistent improvements over different base networks. On ResNet-18 and MobileNet with $\alpha=1.0$, MSNet gains 1.5\% accuracy over both architectures without increasing computations. On the more efficient MobileNet with $\alpha=0.25$, our method gains 3.8\% accuracy with the same amount of computations.
\end{abstract}

\section{Introduction}
\begin{figure*}[t]
\bgroup 
\begin{tabular}{cccc}
\includegraphics[width=0.22\linewidth]{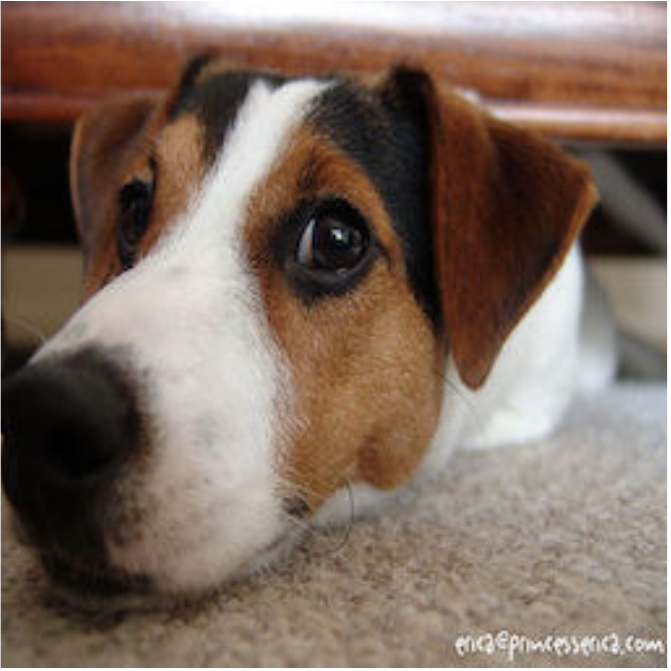} &
\includegraphics[width=0.22\linewidth]{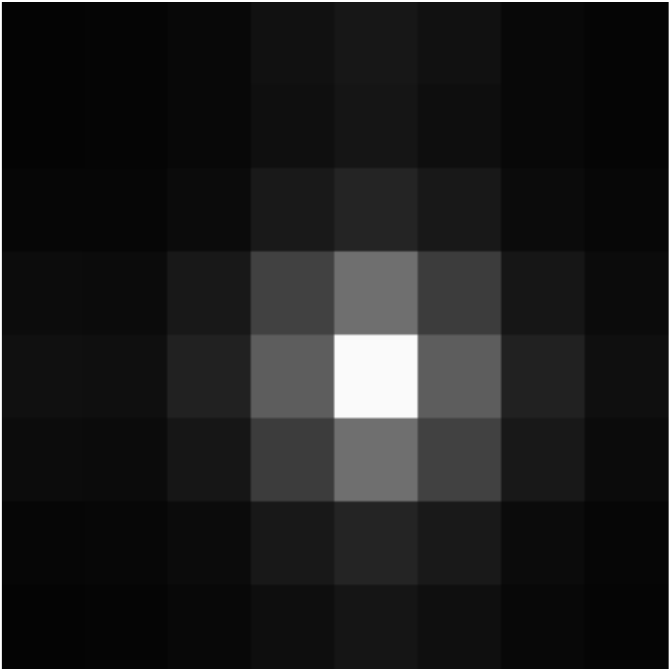} &
\includegraphics[width=0.22\linewidth]{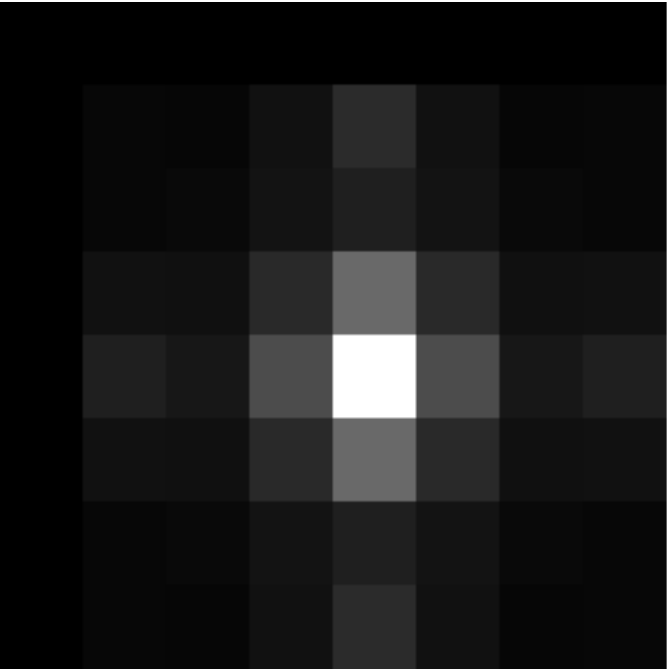} &
\includegraphics[width=0.22\linewidth]{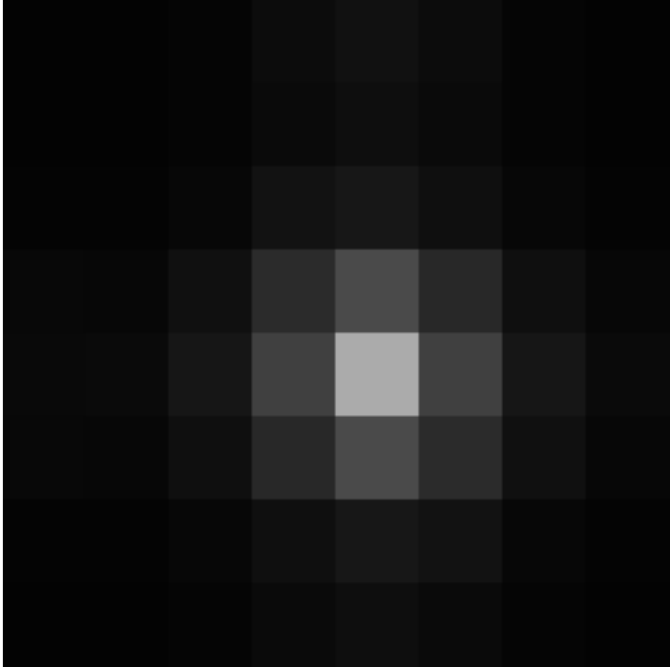} \\
\includegraphics[width=0.22\linewidth]{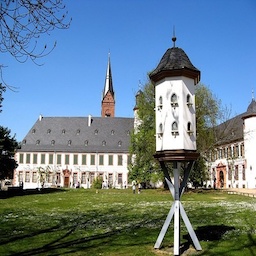} &
\includegraphics[width=0.22\linewidth]{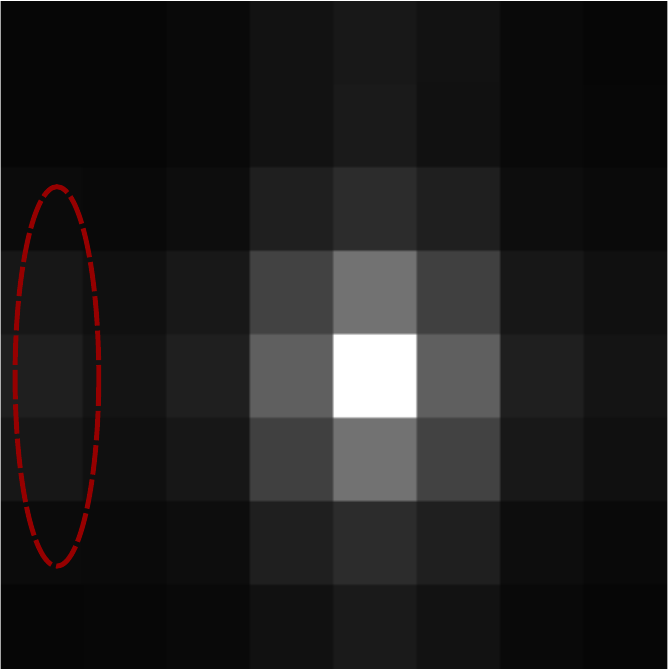} &
\includegraphics[width=0.22\linewidth]{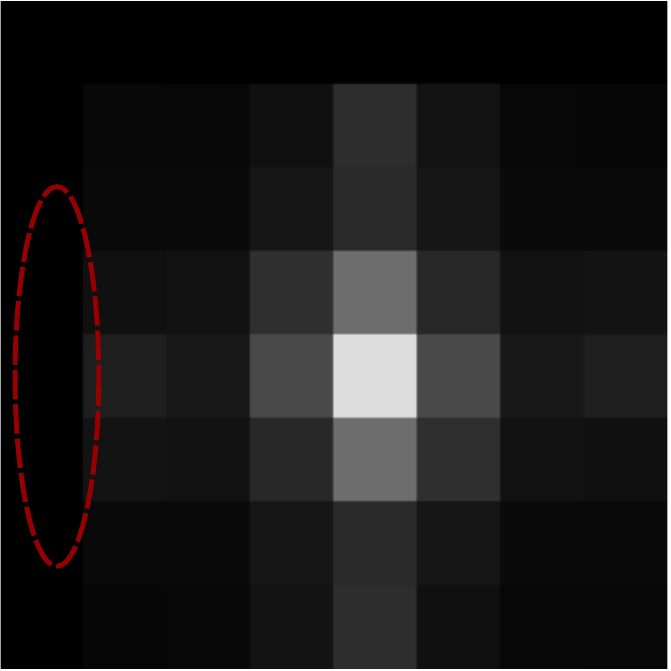} &
\includegraphics[width=0.22\linewidth]{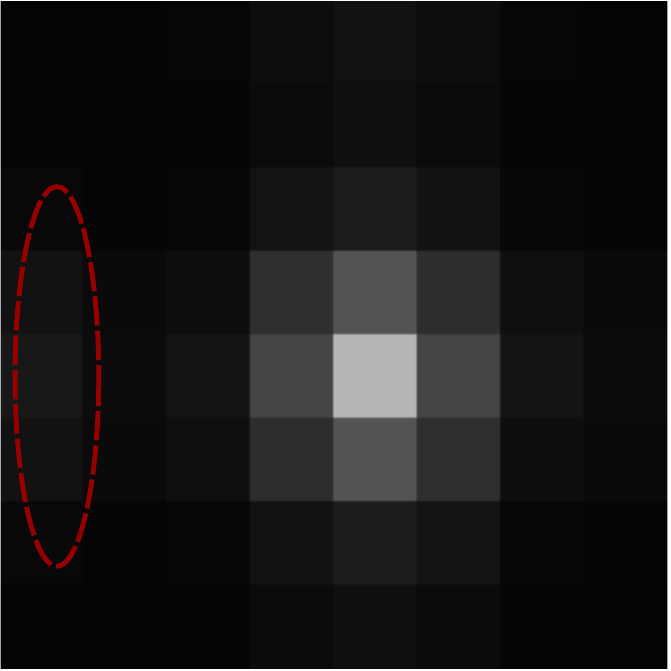} \\
(a) & (b) & (c) & (d)
\end{tabular} \egroup
\caption{Frequency domain analysis. We visualize the magnitude of 2D discrete Fourier transform of final feature map of different networks. We plot the average magnitude of all feature maps from the last convolutional layer. (a): input image to the network. (b): mean magnitude for ResNet-18, it contains mostly the low frequency component. (c): mean magnitude for low resolution network of MS-ResNet-18, compared with ResNet-18, its low frequency component is more concentrated to the DC component. (d): mean magnitude for high resolution network of MS-ResNet-18, compared with ResNet-18 and its low resolution counterpart, it has far less low frequency component. The red ellipse partially covers the high frequency region. From ResNet-18, it shows strong high frequency magnitude. From lower resolution network of MS-ResNet-18, the covered frequency vanishes; while the higher resolution network recovers the high frequency. }
\label{fig:fft}
\end{figure*}
Convolutional neural networks (CNNs) have become the dominant machine learning approach in solving computer vision problems such as, image classification \cite{krizhevsky2012imagenet,simonyan2014very,szegedygoing,he2016deep,cheng2018pretraining,cheng2018revisit}, object detection \cite{girshick2014rich,girshick2015fast,ren2015faster,cheng2018revisiting,wei2018ts2c,cheng18decoupled}, semantic segmentation \cite{long2015fully,chen2018deeplab,zhao2017pyramid,chen2017deeplabv3,deeplabv3plus2018}, etc. In the past several years, great progresses have been achieved in the study of developing large and computationally intensive networks \cite{simonyan2014very,he2016deep,szegedy2017inception} to achieve higher accuracy with the cost of sacrificing speed and efficiency.

Meanwhile, there have been increasing needs of deploying convolutional neural networks into mobile devices with limited computational resources. To address the computational efficiency problem, works \cite{howard2017mobilenets,zhang2017shufflenet,sandler2018inverted} have been proposed to discover more efficient network structure. Common strategies such as replacing convolution operation with more efficient group or depth-wise separable convolutions, using less channels and using lower resolution input have been used.

\begin{figure*}[t]
	\centering
	\includegraphics[width=0.75\textwidth]{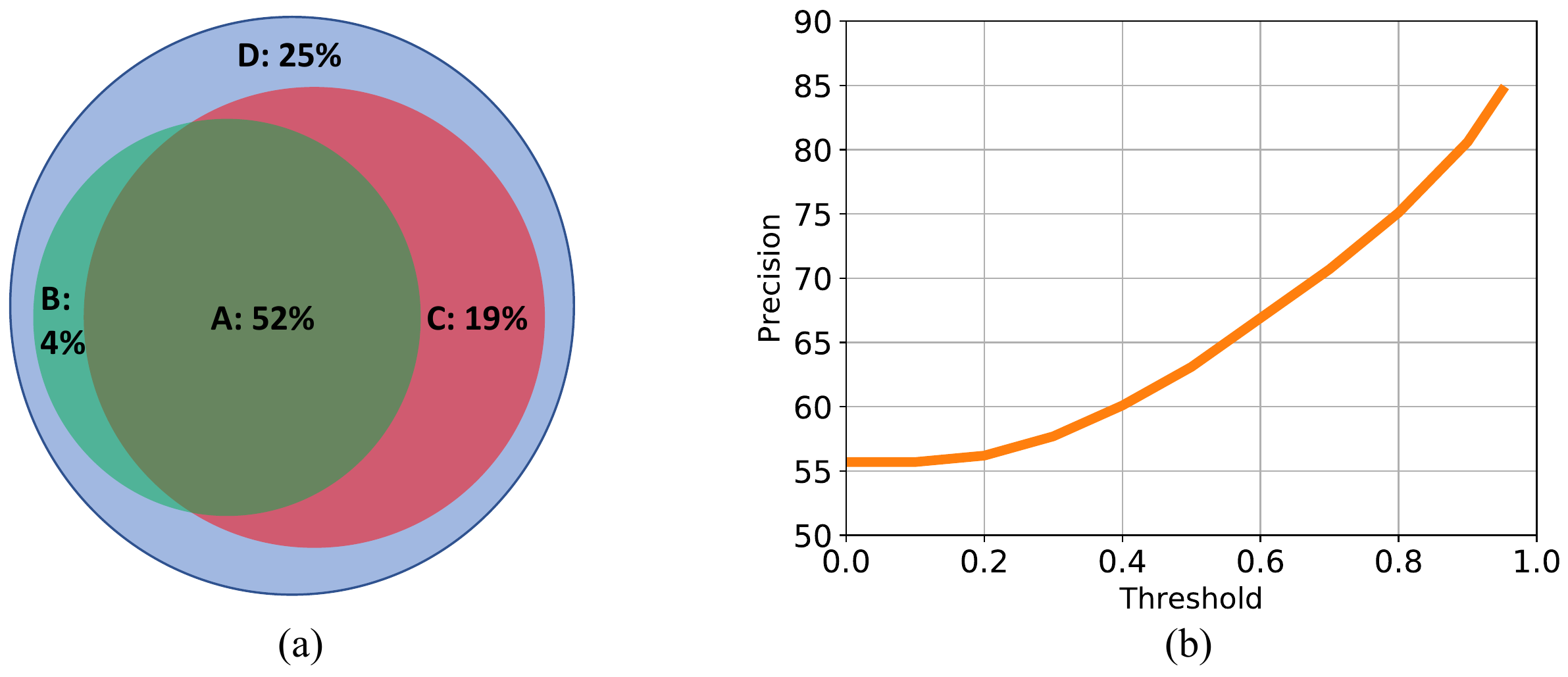}
    \caption{Multi-resolution inference results. The model is ResNet-18. (a): accuracy distribution from the input of $128\times 128$ and $256 \times 256$. 
    Region A: correctly predicted by both $128\times 128$ and $256\times 256$; 
    Region B: correctly predicted by $128 \times 128$ but incorrectly predicted by $256\times 256$; 
    Region C: incorrectly predicted by $128 \times 128$ but correctly predicted by $256\times 256$; 
    Region D: incorrectly predicted by both $128 \times 128$ and $256\times 256$. 
    (b): precision vs thresholds. The threshold is used to select images to be predicted using low resolution input ($128 \times 128$), precision is the accuracy of the selected image. The higher the threshold, the less images are selected and the accuracy becomes higher. 
    }
	\label{fig:analysis}
\end{figure*}

Among these strategies, reducing the input resolution by half is probably the simplest method with the advantage that it can be directly applied during inference. Without the need of re-training, it can reduce the computation cost quadratically. However, while it decreases the computational cost, it usually scarifies the overall accuracy. One question we want to ask and address is: can we do inference on low resolution images without losing accuracy? The answer is yes for some images but not for all the images. Recent works \cite{wang2016studying,cheng2017robust,liu2017enhance,cheng2018visual} propose to directly inference from low resolution input by integrating a super-resolution network with a classification network, but the gain is still marginal. We observe that there are some easy images that can be predicted correctly using both their low resolution and high resolution versions (Figure \ref{fig:analysis} (a)). A straightforward solution is to predict a low resolution image first. If the prediction is incorrect, we make another prediction using the high resolution image. This straightforward solution has a problem that we do not know when classifier fails on low resolution. However, based on the property of softmax function, we can make an assumption that when the prediction of a classifier has a high softmax score, it is more likely to be a correct prediction (Figure \ref{fig:analysis} (b)). Based on this assumption, we can develop a classifier calibration module: 
we first run prediction on low resolution images, 
if the softmax score of the prediction is higher than a threshold, we keep this prediction; if the softmax score of the prediction is lower than the threshold, we use the prediction on the high resolution image. Classifier calibration can be used dynamically during inference to save the average computation.



However, classifier calibration only solves the problem partially. While for an easy image we can use the low resolution predictor to save computational cost quadratically, for a hard image we need to use two predictors corresponding to low and high resolutions as we need the prediction score from the low resolution predictor to tell whether it is an easy or hard image. Checking this process closer, we can easily find that this naive solution does not reuse features of low resolution images anymore once it finishes the prediction on low resolution images, which leads to a waste of computation.

Therefore, the second question we want to ask and address is more intriguing: can low resolution features help high resolution predictions? This question is more interesting because reusing low resolution features will save more computational overheads.


Inspired by the idea of wavelet transform and residual learning, we propose a novel network structure named High Frequency Residual Multi-Scale Network (MSNet) aiming at learning low frequency components and high frequency residuals separately. Note that a feature map normally contains both low frequency components and high frequency components (Figure \ref{fig:fft} (b)). If we upsample a low resolution feature map, it contains mostly low frequency components (Figure \ref{fig:fft} (c)). We find that by reusing the upsampled low resolution features, the network has the ability to learn high frequency residuals with less low frequency components (Figure \ref{fig:fft} (d)). Formally, denoting the desired high resolution feature map by $\ve{y}_{H}$, we hope a lower scale network (a network takes as input low resolution images) learns mainly low frequency components $\ve{y}_{L} = \mathcal{L}(\ve{x}_{L})$ and a higher scale network (a network takes as input high resolution images) learns mainly high frequency residuals $\mathcal{H}(\ve{x}_{H})$. Then the combination of the upsampled low frequency components and high frequency residuals is the desired complete feature map $\ve{y}_{H} = \mathcal{H}(\ve{x}_{H}) + u(\ve{y}_{L})$. Combining \textit{classifier calibration} and \textit{high frequency residual learning}, we achieve a much better computation and accuracy trade-off on multiple efficient networks.

Our contributions are threefold. 1) We propose the MSNet based on high frequency residual learning to efficiently reuse multi-scale features.
2) We propose a classifier calibration module that can dynamically allocate computations during inference. 3)
Our proposed network achieves consistent gains over various architectures without increasing the amount of the computation.

\section{Related Works}

\noindent{\bf Efficient Networks.}
Much attention has been placed on efficient network design recently. Most common methods for designing efficient network or efficient inference are: to use more efficient model component, \eg group convolutions or even depth-wise convolutions \cite{howard2017mobilenets,zhang2017shufflenet,sandler2018inverted}; to perform model compression to reduce parameters and computations \cite{iandola2016squeezenet,yu2018slimmable,yu2019universally}; to use quantization to reduce float32 operations to float16, integer or even binary operations \cite{rastegari2016xnor}; to use knowledge distillation to transfer knowledge of a large network or an ensemble of networks to a small network \cite{hinton2015distilling,romero2014fitnets,polino2018model}. In this work, we focus on the efficient network learning, especially, we target at reducing the computation by reducing the input resolution but preserve the accuracy at the same time. The other model compression/acceleration methods are orthogonal to our method and can further boost the performance.

\noindent{\bf Residual Learning.}
The idea of ResNet \cite{he2016deep} is to add the bypass connection between components
in a feed-forward network. 
It makes deeper networks easier to optimize and has been applied in a wide variety of works \cite{rastegari2016xnor,chollet2016xception,szegedy2017inception,xie2017aggregated}. \cite{chen2017dual} shows the equivalence of residual networks \cite{he2016deep} and densely connected networks \cite{huang2017densely} and the authors propose a general form of residual learning with enhanced performance. 
Compared with ResNet, we explicitly focus on learning the 
high frequency residual information in a multi-scale network architecture, which allows us
to easily allocate the computational cost to trade-off the accuracy. 
Another perspective is that ResNet learns residual in the image domain while our method targets to learn residual in the frequency domain.


\noindent{\bf Cascade Classifiers.}
Cascade classifiers have been widely use for efficient inference. The high level idea of cascade classifiers to classify an image with a sequence of classifiers where the earlier classifiers reject easier images which is called an ``early exit''. The Viola Jones Algorithm \cite{viola2004robust} for face detection uses a hard cascade by Adaboost \cite{freund1997decision}, where multiple week classifiers take different features in a cascade manner. If any of the classifier in the sequence rejects a region, then the region will be classified as a non-facial region. A soft cascade classifier \cite{bourdev2005robust} builds each weak classifier based on the output of all previous classifiers. In the context of deep learning, multiple works have studied the problem of ``early exit'' in a deep model \cite{teerapittayanon2016branchynet,bolukbasi2017adaptive,huang2017multi}. The idea is to inference using early stage features based on the computation budget, however, the problem is early stage features usually do not have high level semantics. In our work, we use the idea of ``early exit'', but instead of building weak classifiers at early stage features, we build a weak classifier on a low resolution input. In this way, our method achieves a good trade-off between speed and high-level features.

\noindent{\bf Multi-Scale Network Design.}
Network designs, either by human \cite{simonyan2014very,he2016deep,huang2017densely} or by architecture search \cite{zoph2017learning,real2018regularized}, mainly focus on finding more powerful building blocks, \eg residual blocks in \cite{he2016deep}, densely connected blocks in \cite{huang2017densely} and machine searched blocks in \cite{zoph2017learning,real2018regularized}. However, all these works use a single resolution input (\eg $224\times224$ or $299\times299$), we find less works study the effect of using features with different resolutions in classification. Recently, there are works \cite{zhang2007real,pedersoli2015coarse,huang2017multi,newell2016stacked,yang2017learning,xie2018interleaved,sun2019deep} focusing on multi-scale design. Although MSDNet \cite{huang2017multi} claims to be ``multi-scale'', we find it is more similar to \cite{saxena2016convolutional} which is a wrapping of different networks into a single network. As far as we know, our work is one of the first to show that reusing low resolution feature helps high resolution prediction. We will give a more detailed comparison in the next section.

\section{High Frequency Residual Learning}

\subsection{Analysis of Multi-Resolution Inference}

Nowadays, most CNN model architecture are fully convolutional with a global average pooling as the last feature layer, making it straightforward to do inference at arbitrary resolution.
Given a well-trained model, we can evaluate its inference accuracy over different scales. Figure \ref{fig:analysis} shows the experiment results. The dataset is ImageNet 2012 \cite{deng2009imagenet}, and the model\footnote{Only for the verifying purpose, we use ResNet-18 \cite{he2016deep} trained with $224 \times 224$ crop. Otherwise, we default the crop size to $256 \times 256$ if not specified explicitly.} is trained and tested on the standard train split and val split, respectively.  
If the test image size is $128 \times 128$, $56\%$ of the images could 
be correctly predicted, within which a majority ($52\% / 56\% = 93\%$) of the images can also be correctly classified if the test size is $256\times 256$. 
Among the incorrectly predicted images by $128 \times 128$, $19\% / 44\%=43\%$
of the images can be corrected by the larger input size of $256 \times 256$. 

In single scale inference, $128 \times 128$ input leads to lower computing cost, 
but only achieves $56\%$ accuracy, while $256 \times 256$ can achieve a higher accuracy but with a penalty of higher cost. If we apply the traditional multi-scale inference, where each image is tested twice, the accuracy could be even higher, but 
leads to a potential waste of computation. For example, $56\%$ of the images can be handled well
by using only smaller scale and testing with both scales gives little benefit (4\%). 

This motivates us to apply the network first onto the small scale to save the 
cost in the target of the $56\%$ correctly predicted images. If the smaller 
scale fails, we resort to the higher scale prediction. Ideally, the upper bound of the accuracy 
is $56\% + 19\% = 75\%$.

The challenging problem is how to properly decide when the lower input size 
fails. In this paper, we adopt a simple strategy to apply a threshold on the 
maximum value of the probability output of the network. The intuition is that if
the network is confident on the prediction (the probability is larger than 
a pre-defined threshold), the accuracy should also be high, 
which is verified in Figure \ref{fig:analysis} (b), illustrating the monotonic increasing relation of accuracy over the thresholds. 
The accuracy is calculated over the samples whose highest prediction score is larger than the threshold. 
As the threshold increases, the number of predictions decreases but their accuracy increases. 
When the threshold is 0, all the predictions are handled by the lower input scale. If 
the threshold is 1, effectively all the images are evaluated by the larger input scale. 

The next question is how to learn the network such that
the lower scale and the higher scale can be cooperated 
efficiently. A naive solution is to learn two networks
independently. One is for the smaller scale, while 
the other is for the larger scale.
However, this is inferior because the features learned 
for the smaller scale network are discarded in the 
feature learning for the larger scale network. 
Instead of learning two networks independently, we
propose a \textit{high frequency
residual} building block to jointly learn 
a multi-scale network. 



\begin{figure*}[t]
	\centering
	\includegraphics[width=1\textwidth]{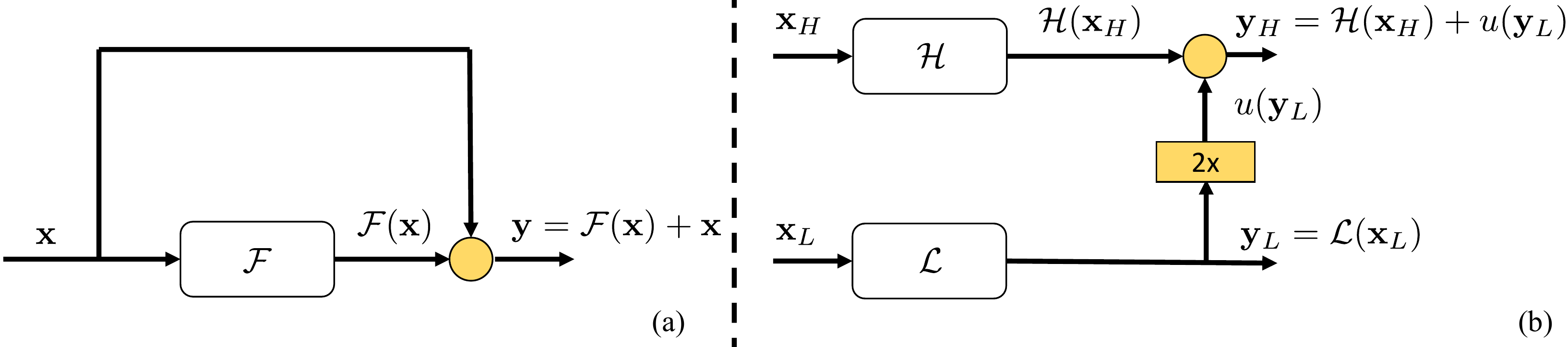}
	\caption{Residual learning block vs. high frequency residual learning block.  (a) a building block for residual learning, $\ve{x}$ is the input feature, $\mathcal{F}(\ve{x})$ is the learned residual and $\ve{y}=\mathcal{F}(\ve{x})+\ve{x}$ is the output feature. (b) a building block for high frequency residual learning in Equation \ref{eq:frequency_residual}.}
	\label{fig:block}
\end{figure*}

\subsection{High Frequency Residual Learning}
\label{subsec:block}
Figure \ref{fig:block} (b) depicts the proposed
building block. 
The input is $\mathbf{x}_L$ and 
$\mathbf{x}_H$ in the real domain 
for lower scale and higher
scale, respectively. 
Passing the two signals each through several 
linear layers interleaved with nonlinear activations, 
we can have
more abstract feature representations of $\mathcal{L}(\mathbf{x}_L)$, and $\mathcal{H}(\mathbf{x}_H)$. 
Instead of learning them separately, we 
enrich the larger scale pass by the upsampled  
lower signal $\mathbf{y}_L = \mathcal{L}(\mathbf{x}_L)$.
That is, the output of the higher scale is 
$\mathbf{y}_H = \mathcal{H}(\mathbf{x}_H) + u(\mathbf{y}_L)$, 
where $u(\cdot)$ is an upsampling function. Formally, the building block can be written as
\begin{align}\label{eq:frequency_residual}
     \ve{y}_{L} & = \mathcal{L}(\ve{x}_{L}) \\
     \ve{y}_{H} & = \mathcal{H}(\ve{x}_{H}) + u(\ve{y}_{L}).
\end{align}

The upsampling operation helps match the spatial resolution 
between two different scales to make the aggregation feasible. 
Meanwhile, the function is a low-pass filter,
which reduces the high-frequency information to pass through. 
For example, if we implement the upsampling by a nearest neighbor 
interpolation, the impulse response is a rectangle function and 
the corresponding frequency function is a sinc function, which allows
more lower frequency to pass and blocks more higher frequency. 

Since the flowing information from the bottom to the top is mostly the low frequency information, the higher scale is expected to focus on
the high frequency information learning. 
This intuition can also be verified in Figure \ref{fig:fft} (d), which contains less lower frequency information than \ref{fig:fft} (b). Thus, we call it \textit{high frequency residual} learning.

Figure \ref{fig:block} (a) shows the residual 
learning \cite{he2016deep}
for comparison. Since we explicitly split the signal 
into different frequency bands, it is easier 
to cooperatively apply the multi-scale inference efficiently. 
Besides, the residual learning block can be 
part of 
the high frequency learning block by inserting it into $\mathcal{H}$ and $\mathcal{L}$.

\begin{figure*}[t]
	\centering
	\includegraphics[width=1\textwidth]{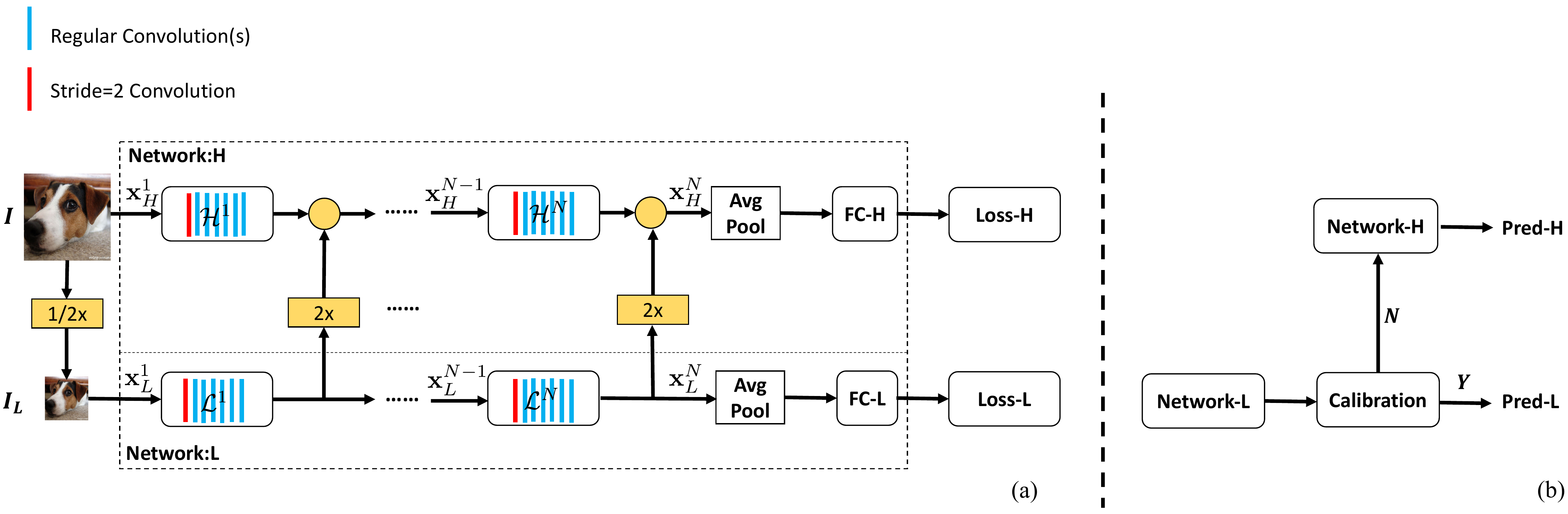}
    \caption{Multi-Scale Network (MSNet) network structure. (a) MSNet structure for training: MSNet is composite of a higher scale network (Network-H in upper dashed box) and a lower scale network (Network-L in lower dashed box). Both networks can make prediction and are trained jointly. (b) MSNet with calibration in inference: if the calibration module produces ``Y'' (softmax score higher then the threshold), MSNet only use Network-L prediction (pred-L); it the calibration module produces ``N'' (softmax score lower then the threshold), MSNet use Network-H prediction (pred-H). Note that when using Network-H, features in Network-L are reused and passed into Network-H together with the input image.}
	\label{fig:network}
\end{figure*}

\subsection{High Frequency Residual Multi-Scale Network}
By stacking multiple high frequency residual learning blocks, we construct a multi-scale network (MSNet) as shown in
Figure \ref{fig:network} (a). 
It consists of a lower scale network, a higher scale network and a calibration module. 
The raw image serves as the input to the higher scale network, while the downsampled version to the smaller scale network. 
We use the superscript to distinguish different blocks. 
Within the $i$-th block, let $\mathbf{x}_L^i$ and 
$\mathbf{x}_H^i$ be the input to the lower scale and higher
scale networks, respectively. 
Within each feature extraction module $\mathcal{L}^{i}$
and $\mathcal{H}^{i}$, we first apply a convolutional layer
with stride as 2 to reduce the feature map size. 
Instead of using $\mathbf{y}$ in Eq. \ref{eq:frequency_residual}, 
we use $\mathbf{x}_L^{i + 1}$
and $\mathbf{x}_H^{i + 1}$ to denote outputs, which are also the 
input of the next block. 
After the last block, a global average pooling layer is added
before we apply a linear layer and the softmax layer to output the classification 
result. 

The inference stage of MSNet is shown in Figure \ref{fig:network} (b), the test image is first downsampled
and then fed into the lower scale network. 
The classification result is sent to the calibration module. 
If the maximum probability is larger than a pre-defined
threshold, the calibration module outputs this result and terminates the inference process. Otherwise, it 
enables the upsampling of low resolution features from the lower scale network
and the forward pass of the higher scale network. In this case, the calibration module takes the output
from the higher scale network as the final prediction. 

During training, we remove the calibration module 
and add the cross entropy loss to each of the classifier. 
Since all the components are differentiable, we can apply
the off-the-shelf gradient descent algorithm to update the parameters. 

\section{Experiments}
\begin{figure*}[t]
	\centering
	\includegraphics[width=1\textwidth]{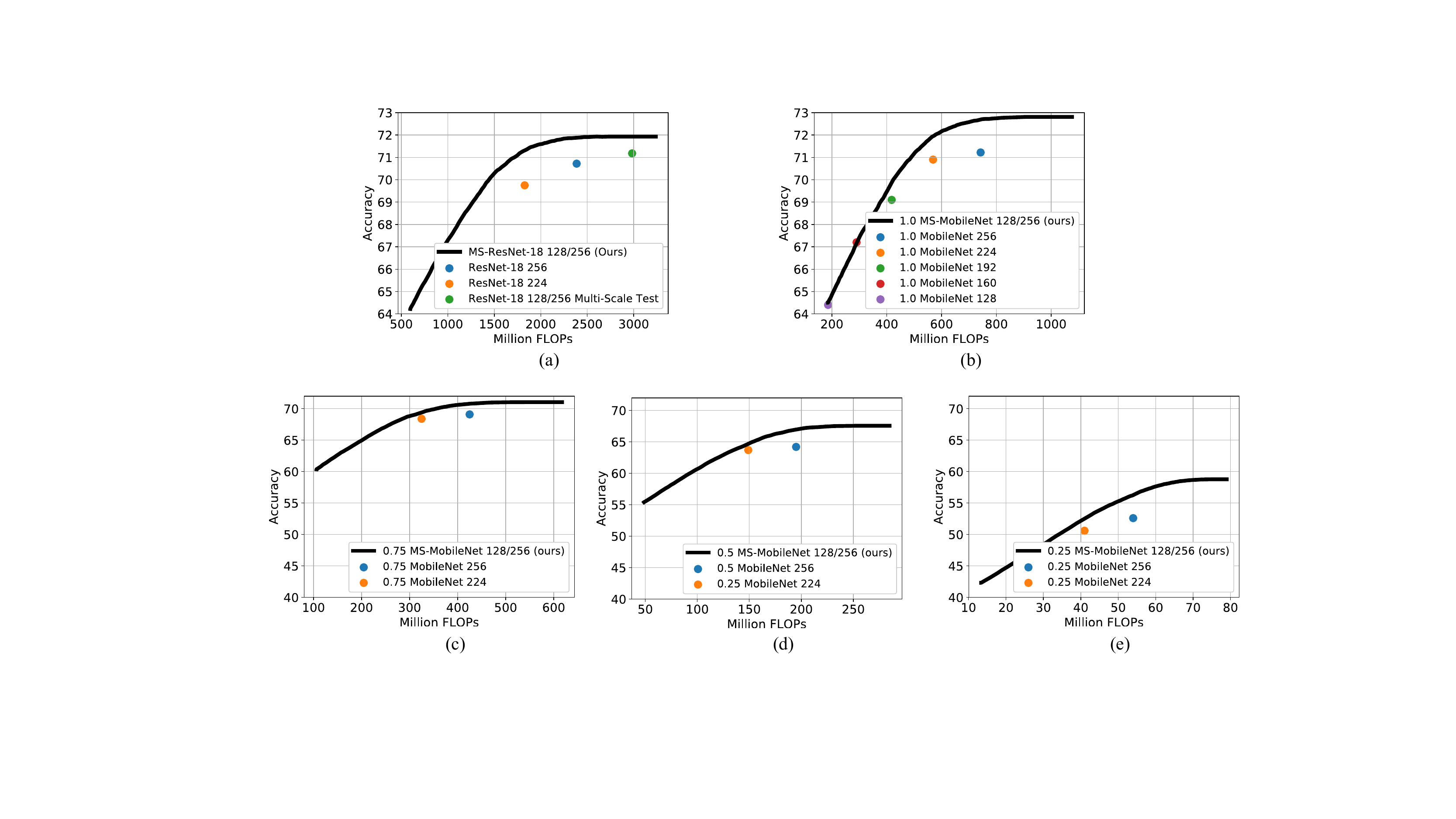}
	\caption{Our MSNet calibration results with different networks. (a) MS-ResNet-18 vs. ResNet-18. (b)-(e) MS-MobileNet vs. MobileNet. The black curves (our results) are obtained by varying threshold values.} 
	\label{fig:result}
    \vspace{-5mm}
\end{figure*}

\subsection{Settings}

We evaluate our method on the widely-used ImageNet 2012 classification dataset \cite{deng2009imagenet}.
All the models are trained on the 1.28 million training images, and we report the top-1 accuracy of a single center crop on the 50k validation set. 
If the crop size is $224 \times 224$, we first resize the shorter side of the image to 256 before doing the center crop. 
If the crop size is $256 \times 256$, we resize the shorter side to 300. We use standard data augmentation in \cite{he2016deep,howard2017mobilenets} and train the model for 120 epochs with initial learning rate 0.1, reduced by 0.1 at 30, 60, 90 epochs.

Given a base model, e.g. ResNet, 
to construct its MSNet version, we extract each stage of the base model
and use it as the feature abstraction part of $\mathcal{H}$
and $\mathcal{L}$. 
We conduct experiments on ResNet-18 and MobileNet, and the corresponding MSNet version is called MS-ResNet-18 and MS-MobileNet, respectively. 
For the calibration module, we use different threshold to trade-off the 
computation and the accuracy. The computation cost is measured by the  FLOPS as in \cite{he2016deep,howard2017mobilenets}. In MSNet, different images could have different FLOPS cost, and we report the average cost.
Without explicitly explaining, the higher scale is $256 \times 256$ and
the lower scale is $128 \times 128$. 


\subsection{Results}
\subsubsection{MS-ResNet-18 vs ResNet-18}
Table \ref{res18} shows the experiment results. 
First, we reproduce the ResNet-18 network and get an accuracy of 69.8\% over the $224\times 224$ input. 
By setting the threshold as 0.48, our MS-ResNet-18 can achieve similar accuracy of 
$69.9\%$, but consumes only $77\%$ ($1413/1827$) FLOPS. If the threshold is 0.66, the 
computation cost is similar, but the accuracy of our MS-ResNet-18 is boosted to $71.3\%$ with an absolute gain of $1.5\%$.

If the input size of ResNet-18 is $256\times 256$, the accuracy is 70.7\%. Still, with similar accuracy, the FLOPS cost can be reduced to $68\%$ ($1625/2386$) with our MS-ResNet-18. 

Next, we perform the traditional multi-scale testing by averaging prediction results of different resolution (128 and 256) inputs for ResNet-18, which achieves 71.2\% accuracy but consumes 25\% more computations. Using our proposed MS-ResNet-18, we achieve 71.9\% accuracy and consumes only $79\%$ ($2358/2983$) computations.
\begin{table}[t]
\parbox{1.0\linewidth}{
\centering
\caption{Experiment results between our MS-ResNet-18 and ResNet-18.}
\label{res18}
  \scriptsize
  \begin{tabular}{lcc}
	\hline
	Method & Million FLOPs & Top-1 Accuracy\\
	\hline
	ResNet-18 (224) &  1827 & 69.8\% \\
    MS-ResNet-18 (128/256), thresh=0.48 & 1413 & 69.9\% \\
    MS-ResNet-18 (128/256), thresh=0.66 & 1811 & 71.3\% \\
    \hline
    ResNet-18 (256) &  2386 & 70.7\% \\
    MS-ResNet-18 (128/256), thresh=0.57 & 1625 & 70.7\% \\
    ResNet-18 (multi-scale test) &  2983 & 71.2\%  \\
    MS-ResNet-18 (128/256), thresh=0.90 & 2358 & 71.9\% \\
	\hline
  \end{tabular}
}
\end{table}

\begin{table*}[t]
\parbox{1\linewidth}{
\centering
  \scriptsize
  \caption{Experiment results between our MS-MobileNet and MobileNet.}
\label{mobile}
  \begin{tabular}{lccc}
	\hline
	Method & Million FLOPs & Top-1 Accuracy & Accuracy Gain\\
	\hline
	1.0 MobileNet (128) &  186 & 64.4\% \\
    1.0 MS-MobileNet (128/256), thresh=0.06&  186 & 64.5\% & +0.1\% \\
    1.0 MobileNet (192) &  418 & 69.1\% \\
    1.0 MS-MobileNet (128/256), thresh=0.46 &  417 & 69.8\% &+0.7\%\\
    1.0 MobileNet (224) &  569 & 70.9\% \\
    1.0 MS-MobileNet (128/256), thresh=0.67 &  566 & 71.9\% &+1.0\%\\
    1.0 MobileNet (256) &  743 & 71.2\% \\
    1.0 MS-MobileNet (128/256), thresh=0.90 &  739 & 72.7\% &+1.5\%\\
    \hline
    0.75 MobileNet (224) &  325 & 68.4\% \\
    0.75 MS-MobileNet (128/256), thresh=0.57 &  321 & 69.3\% &+0.9\% \\
    0.75 MobileNet (256) &  425 & 69.1\% \\
    0.75 MS-MobileNet (128/256), thresh=0.83 &  425 & 70.8\% &+1.7\% \\
    \hline
    0.5 MobileNet (224) &  149 & 63.7\% \\
    0.5 MS-MobileNet (128/256), thresh=0.47 &  148 & 64.6\% &+0.9\%\\
    0.5 MobileNet (256) &  195 & 64.2\% \\
    0.5 MS-MobileNet (128/256), thresh=0.71 &  195 & 67.0\% &+2.8\% \\
    \hline
    0.25 MobileNet (224) &  41 & 50.6\% \\
    0.25 MS-MobileNet (128/256), thresh=0.28 &  41 & 52.5\% &+1.9\% \\
    0.25 MobileNet (256) &  54 & 52.6\% \\
    0.25 MS-MobileNet (128/256), thresh=0.46 &  54 & 56.4\% &+3.8\%\\
    \hline
  \end{tabular}
}
\end{table*}

\subsubsection{MS-MobileNet vs MobileNet}
Following~\cite{howard2017mobilenets}, we shrink the number of channels by a 
factor of $\alpha$ to achieves different trade-offs on the MobileNet. 
The corresponding MSNet version is also shrunk accordingly. 
We pre-fix the network name with $\alpha$ in the experiment results, shown in Table \ref{mobile}. 
For MobileNet with $\alpha=1.0$ where $\alpha$ is the width multiplier to thin a network uniformly at each layer, we change the input scale to obtain different accuracies. 
As the input resolution increases, our MS-MobileNet shows larger advantages under the same computational cost over the MobileNet. This clearly demonstrates that the lower scale in our MS-MobileNet could identify correctly a certain amount of images without resorting to the higher scale network.  

When 
$\alpha$ becomes smaller, we observe that our MS-MobileNet achieves 
even higher accuracy relatively under the same amount of computations, e.g. comparing the MobileNet with $256\times 256$ input, the gain is 1.7\% for $\alpha=0.75$, 2.8\% for $\alpha=0.5$ and 3.8\% for $\alpha=0.25$.
This clearly demonstrates our approach shows superior advantages for small models and promising applications in mobile and embedded devices.

\subsubsection{MS-DenseNet-121 vs DenseNet-121}
When applying our method to DenseNet, we use crop $224 \times 224$ as input and use cropping or zero-padding to align upsampled low resolution feature maps to the corresponding high resolution feature maps if their spatial resolution does not match exactly.

Table \ref{dense121} shows our experimental results on DenseNet-121. The original DenseNet-121 has an accuracy of 75\% with 2.8 Billion floating point operations. By setting the threshold to 0.6, out MS-DenseNet-121 achieves similar accuracy (75\%) but with less computation (1.7 Billion FLOPs). That is, when combined with DenseNet, our method is on par with  state-of-the-art method \cite{huang2017multi}. 
With the same computation (2.8 BFLOPs), our method achieves a much higher accuracy (76.1\%) than the original DenseNet with a margin of 1.1\%.

\begin{table}[t]
\parbox{1\linewidth}{
\centering
\caption{Experiment results between our MS-DenseNet-121 and DenseNet-121.}
\label{dense121}
  \scriptsize
  \begin{tabular}{lcc}
	\hline
	Method & Billion FLOPs & Top-1 Accuracy\\
	\hline
	DenseNet-121 (224) &  2.8 & 75.0\% \\
    MSDNet \cite{huang2017multi} & 1.7 & 75.0\% \\
    MS-DenseNet-121 (112/224), thresh=0.60 & 1.7 & 75.0\% \\
    MS-DenseNet-121 (112/224), thresh=0.95 & 2.8 & 76.1\% \\
	\hline
  \end{tabular}
}
\end{table}

\subsection{Speed-Accuracy Trade-off}
Since different thresholds lead to different computation costs and  accuracies, we can achieve a speed-accuracy trade-off by varying the threshold values. 
We illustrate the relationship by enumerating multiple thresholds in 
Figure \ref{fig:result} for MS-ResNet-18 and MS-MobileNet (the black curve). 
The baseline approaches (colored dots)
are also shown in the figure. 
We can easily observe the advantages of the MSNet over the baselines. 


\section{Conclusions}

In this paper, we have presented a novel high frequency residual learning framework that decouples the learning of low frequency feature and high frequency feature. We demonstrate that computations can be saved by using a low resolution network to approximate the process of learning low frequency features. We have also proposed a classifier calibration module which can dynamically allocate computation resources during inference and lead to a better speed and accuracy trade-off. In our future work, we will design more flexible high frequency residual networks that can take as input arbitrary size of images. We will also demonstrate the effectiveness of high frequency residual learning in other recognition task, \eg object detection and semantic segmentation, in the future.

\paragraph{Acknowledgements.} Bowen Cheng and Thomas Huang are in part supported by IBM-Illinois Center for Cognitive Computing Systems Research (C3SR) - a research collaboration as part of the IBM AI Horizons Network. The authors thank Bin Xiao for helpful discussions.

\bibliography{egbib}

\begin{thebibliography}{52}
\providecommand{\natexlab}[1]{#1}
\providecommand{\url}[1]{\texttt{#1}}
\expandafter\ifx\csname urlstyle\endcsname\relax
  \providecommand{\doi}[1]{doi: #1}\else
  \providecommand{\doi}{doi: \begingroup \urlstyle{rm}\Url}\fi

\bibitem[Bolukbasi et~al.(2017)Bolukbasi, Wang, Dekel, and
  Saligrama]{bolukbasi2017adaptive}
Tolga Bolukbasi, Joseph Wang, Ofer Dekel, and Venkatesh Saligrama.
\newblock Adaptive neural networks for efficient inference.
\newblock \emph{{ICML}}, 2017.

\bibitem[Bourdev and Brandt(2005)]{bourdev2005robust}
Lubomir Bourdev and Jonathan Brandt.
\newblock Robust object detection via soft cascade.
\newblock In \emph{{IEEE CVPR}}, volume~2, pages 236--243. IEEE, 2005.

\bibitem[Chen et~al.(2017{\natexlab{a}})Chen, Papandreou, Schroff, and
  Adam]{chen2017deeplabv3}
Liang-Chieh Chen, George Papandreou, Florian Schroff, and Hartwig Adam.
\newblock Rethinking atrous convolution for semantic image segmentation.
\newblock \emph{arXiv preprint arXiv:1706.05587}, 2017{\natexlab{a}}.

\bibitem[Chen et~al.(2018{\natexlab{a}})Chen, Papandreou, Kokkinos, Murphy, and
  Yuille]{chen2018deeplab}
Liang-Chieh Chen, George Papandreou, Iasonas Kokkinos, Kevin Murphy, and Alan~L
  Yuille.
\newblock Deeplab: Semantic image segmentation with deep convolutional nets,
  atrous convolution, and fully connected crfs.
\newblock \emph{{IEEE TPAMI}}, 40\penalty0 (4):\penalty0 834--848,
  2018{\natexlab{a}}.

\bibitem[Chen et~al.(2018{\natexlab{b}})Chen, Zhu, Papandreou, Schroff, and
  Adam]{deeplabv3plus2018}
Liang-Chieh Chen, Yukun Zhu, George Papandreou, Florian Schroff, and Hartwig
  Adam.
\newblock Encoder-decoder with atrous separable convolution for semantic image
  segmentation.
\newblock In \emph{{ECCV}}, 2018{\natexlab{b}}.

\bibitem[Chen et~al.(2017{\natexlab{b}})Chen, Li, Xiao, Jin, Yan, and
  Feng]{chen2017dual}
Yunpeng Chen, Jianan Li, Huaxin Xiao, Xiaojie Jin, Shuicheng Yan, and Jiashi
  Feng.
\newblock Dual path networks.
\newblock In \emph{{NIPS}}, pages 4470--4478, 2017{\natexlab{b}}.

\bibitem[Cheng et~al.(2017)Cheng, Wang, Zhang, Li, Liu, Yang, Huang, and
  Huang]{cheng2017robust}
Bowen Cheng, Zhangyang Wang, Zhaobin Zhang, Zhu Li, Ding Liu, Jianchao Yang,
  Shuai Huang, and Thomas~S Huang.
\newblock Robust emotion recognition from low quality and low bit rate video: A
  deep learning approach.
\newblock In \emph{2017 Seventh International Conference on Affective Computing
  and Intelligent Interaction (ACII)}, pages 65--70. IEEE, 2017.

\bibitem[Cheng et~al.(2018{\natexlab{a}})Cheng, Liu, Wang, Zhang, and
  Huang]{cheng2018visual}
Bowen Cheng, Ding Liu, Zhangyang Wang, Haichao Zhang, and Thomas~S Huang.
\newblock Visual recognition in very low-quality settings: Delving into the
  power of pre-training.
\newblock In \emph{Thirty-Second AAAI Conference on Artificial Intelligence},
  2018{\natexlab{a}}.

\bibitem[Cheng et~al.(2018{\natexlab{b}})Cheng, Wei, Shi, Chang, Xiong, and
  Huang]{cheng2018pretraining}
Bowen Cheng, Yunchao Wei, Honghui Shi, Shiyu Chang, Jinjun Xiong, and Thomas~S
  Huang.
\newblock Revisiting pre-training: An efficient training method for image
  classification.
\newblock \emph{arXiv preprint arXiv:1811.09347}, 2018{\natexlab{b}}.

\bibitem[Cheng et~al.(2018{\natexlab{c}})Cheng, Wei, Shi, Feris, Xiong, and
  Huang]{cheng18decoupled}
Bowen Cheng, Yunchao Wei, Honghui Shi, Rogerio Feris, Jinjun Xiong, and Thomas
  Huang.
\newblock Decoupled classification refinement: Hard false positive suppression
  for object detection.
\newblock \emph{arXiv preprint arXiv:1810.04002}, 2018{\natexlab{c}}.

\bibitem[Cheng et~al.(2018{\natexlab{d}})Cheng, Wei, Shi, Feris, Xiong, and
  Huang]{cheng2018revisiting}
Bowen Cheng, Yunchao Wei, Honghui Shi, Rogerio Feris, Jinjun Xiong, and Thomas
  Huang.
\newblock Revisiting rcnn: On awakening the classification power of faster
  rcnn.
\newblock In \emph{{ECCV}}, 2018{\natexlab{d}}.

\bibitem[Cheng et~al.(2018{\natexlab{e}})Cheng, Xiao, Guo, Hu, Wang, and
  Zhang]{cheng2018revisit}
Bowen Cheng, Rong Xiao, Yandong Guo, Yuxiao Hu, Jianfeng Wang, and Lei Zhang.
\newblock Revisit multinomial logistic regression in deep learning: Data
  dependent model initialization for image recognition.
\newblock \emph{arXiv preprint arXiv:1809.06131}, 2018{\natexlab{e}}.

\bibitem[Chollet(2016)]{chollet2016xception}
Fran{\c{c}}ois Chollet.
\newblock Xception: Deep learning with depthwise separable convolutions.
\newblock \emph{arXiv preprint}, 2016.

\bibitem[Deng et~al.(2009)Deng, Dong, Socher, Li, Li, and
  Fei-Fei]{deng2009imagenet}
Jia Deng, Wei Dong, Richard Socher, Li-Jia Li, Kai Li, and Li~Fei-Fei.
\newblock Imagenet: A large-scale hierarchical image database.
\newblock In \emph{{IEEE CVPR}}, pages 248--255. IEEE, 2009.

\bibitem[Freund and Schapire(1997)]{freund1997decision}
Yoav Freund and Robert~E Schapire.
\newblock A decision-theoretic generalization of on-line learning and an
  application to boosting.
\newblock \emph{Journal of computer and system sciences}, 55\penalty0
  (1):\penalty0 119--139, 1997.

\bibitem[Girshick(2015)]{girshick2015fast}
Ross Girshick.
\newblock Fast r-cnn.
\newblock In \emph{{IEEE ICCV}}, pages 1440--1448. IEEE, 2015.

\bibitem[Girshick et~al.(2014)Girshick, Donahue, Darrell, and
  Malik]{girshick2014rich}
Ross Girshick, Jeff Donahue, Trevor Darrell, and Jitendra Malik.
\newblock Rich feature hierarchies for accurate object detection and semantic
  segmentation.
\newblock In \emph{{IEEE CVPR}}, pages 580--587, 2014.

\bibitem[He et~al.(2016)He, Zhang, Ren, and Sun]{he2016deep}
Kaiming He, Xiangyu Zhang, Shaoqing Ren, and Jian Sun.
\newblock Deep residual learning for image recognition.
\newblock In \emph{{IEEE CVPR}}, pages 770--778, 2016.

\bibitem[Hinton et~al.(2015)Hinton, Vinyals, and Dean]{hinton2015distilling}
Geoffrey Hinton, Oriol Vinyals, and Jeff Dean.
\newblock Distilling the knowledge in a neural network.
\newblock \emph{arXiv preprint arXiv:1503.02531}, 2015.

\bibitem[Howard et~al.(2017)Howard, Zhu, Chen, Kalenichenko, Wang, Weyand,
  Andreetto, and Adam]{howard2017mobilenets}
Andrew~G Howard, Menglong Zhu, Bo~Chen, Dmitry Kalenichenko, Weijun Wang,
  Tobias Weyand, Marco Andreetto, and Hartwig Adam.
\newblock Mobilenets: Efficient convolutional neural networks for mobile vision
  applications.
\newblock \emph{arXiv preprint arXiv:1704.04861}, 2017.

\bibitem[Huang et~al.(2017)Huang, Liu, Weinberger, and van~der
  Maaten]{huang2017densely}
Gao Huang, Zhuang Liu, Kilian~Q Weinberger, and Laurens van~der Maaten.
\newblock Densely connected convolutional networks.
\newblock In \emph{{IEEE CVPR}}, volume~1, page~3, 2017.

\bibitem[Huang et~al.(2018)Huang, Chen, Li, Wu, van~der Maaten, and
  Weinberger]{huang2017multi}
Gao Huang, Danlu Chen, Tianhong Li, Felix Wu, Laurens van~der Maaten, and
  Kilian~Q Weinberger.
\newblock Multi-scale dense networks for resource efficient image
  classification.
\newblock \emph{ICLR}, 2018.

\bibitem[Iandola et~al.(2016)Iandola, Han, Moskewicz, Ashraf, Dally, and
  Keutzer]{iandola2016squeezenet}
Forrest~N Iandola, Song Han, Matthew~W Moskewicz, Khalid Ashraf, William~J
  Dally, and Kurt Keutzer.
\newblock Squeezenet: Alexnet-level accuracy with 50x fewer parameters and< 0.5
  mb model size.
\newblock \emph{arXiv preprint arXiv:1602.07360}, 2016.

\bibitem[Krizhevsky et~al.(2012)Krizhevsky, Sutskever, and
  Hinton]{krizhevsky2012imagenet}
Alex Krizhevsky, Ilya Sutskever, and Geoffrey~E Hinton.
\newblock Imagenet classification with deep convolutional neural networks.
\newblock In \emph{{NIPS}}, pages 1097--1105, 2012.

\bibitem[Liu et~al.(2017)Liu, Cheng, Wang, Zhang, and Huang]{liu2017enhance}
Ding Liu, Bowen Cheng, Zhangyang Wang, Haichao Zhang, and Thomas~S Huang.
\newblock Enhance visual recognition under adverse conditions via deep
  networks.
\newblock \emph{arXiv preprint arXiv:1712.07732}, 2017.

\bibitem[Long et~al.(2015)Long, Shelhamer, and Darrell]{long2015fully}
Jonathan Long, Evan Shelhamer, and Trevor Darrell.
\newblock Fully convolutional networks for semantic segmentation.
\newblock In \emph{{IEEE CVPR}}, pages 3431--3440, 2015.

\bibitem[Newell et~al.(2016)Newell, Yang, and Deng]{newell2016stacked}
Alejandro Newell, Kaiyu Yang, and Jia Deng.
\newblock Stacked hourglass networks for human pose estimation.
\newblock In \emph{European Conference on Computer Vision}, pages 483--499.
  Springer, 2016.

\bibitem[Pedersoli et~al.(2015)Pedersoli, Vedaldi, Gonzalez, and
  Roca]{pedersoli2015coarse}
Marco Pedersoli, Andrea Vedaldi, Jordi Gonzalez, and Xavier Roca.
\newblock A coarse-to-fine approach for fast deformable object detection.
\newblock \emph{Pattern Recognition}, 48\penalty0 (5):\penalty0 1844--1853,
  2015.

\bibitem[Polino et~al.(2018)Polino, Pascanu, and Alistarh]{polino2018model}
Antonio Polino, Razvan Pascanu, and Dan Alistarh.
\newblock Model compression via distillation and quantization.
\newblock \emph{arXiv preprint arXiv:1802.05668}, 2018.

\bibitem[Rastegari et~al.(2016)Rastegari, Ordonez, Redmon, and
  Farhadi]{rastegari2016xnor}
Mohammad Rastegari, Vicente Ordonez, Joseph Redmon, and Ali Farhadi.
\newblock Xnor-net: Imagenet classification using binary convolutional neural
  networks.
\newblock In \emph{{ECCV}}, pages 525--542. Springer, 2016.

\bibitem[Real et~al.(2018)Real, Aggarwal, Huang, and Le]{real2018regularized}
Esteban Real, Alok Aggarwal, Yanping Huang, and Quoc~V Le.
\newblock Regularized evolution for image classifier architecture search.
\newblock 2018.

\bibitem[Ren et~al.(2015)Ren, He, Girshick, and Sun]{ren2015faster}
Shaoqing Ren, Kaiming He, Ross Girshick, and Jian Sun.
\newblock Faster r-cnn: Towards real-time object detection with region proposal
  networks.
\newblock In \emph{{NIPS}}, pages 91--99, 2015.

\bibitem[Romero et~al.(2014)Romero, Ballas, Kahou, Chassang, Gatta, and
  Bengio]{romero2014fitnets}
Adriana Romero, Nicolas Ballas, Samira~Ebrahimi Kahou, Antoine Chassang, Carlo
  Gatta, and Yoshua Bengio.
\newblock Fitnets: Hints for thin deep nets.
\newblock \emph{arXiv preprint arXiv:1412.6550}, 2014.

\bibitem[Sandler et~al.(2018)Sandler, Howard, Zhu, Zhmoginov, and
  Chen]{sandler2018inverted}
Mark Sandler, Andrew Howard, Menglong Zhu, Andrey Zhmoginov, and Liang-Chieh
  Chen.
\newblock Inverted residuals and linear bottlenecks: Mobile networks for
  classification, detection and segmentation.
\newblock \emph{arXiv preprint arXiv:1801.04381}, 2018.

\bibitem[Saxena and Verbeek(2016)]{saxena2016convolutional}
Shreyas Saxena and Jakob Verbeek.
\newblock Convolutional neural fabrics.
\newblock In \emph{Advances in Neural Information Processing Systems}, pages
  4053--4061, 2016.

\bibitem[Simonyan and Zisserman(2014)]{simonyan2014very}
Karen Simonyan and Andrew Zisserman.
\newblock Very deep convolutional networks for large-scale image recognition.
\newblock \emph{arXiv preprint arXiv:1409.1556}, 2014.

\bibitem[Sun et~al.(2019)Sun, Xiao, Liu, and Wang]{sun2019deep}
Ke~Sun, Bin Xiao, Dong Liu, and Jingdong Wang.
\newblock Deep high-resolution representation learning for human pose
  estimation.
\newblock \emph{arXiv preprint arXiv:1902.09212}, 2019.

\bibitem[Szegedy et~al.()Szegedy, Liu, Jia, Sermanet, Reed, Anguelov, Erhan,
  Vanhoucke, Rabinovich, Rick~Chang, et~al.]{szegedygoing}
Christian Szegedy, Wei Liu, Yangqing Jia, Pierre Sermanet, Scott Reed, Dragomir
  Anguelov, Dumitru Erhan, Vincent Vanhoucke, Andrew Rabinovich, Jen-Hao
  Rick~Chang, et~al.
\newblock Going deeper with convolutions.
\newblock In \emph{{IEEE CVPR}}.

\bibitem[Szegedy et~al.(2017)Szegedy, Ioffe, Vanhoucke, and
  Alemi]{szegedy2017inception}
Christian Szegedy, Sergey Ioffe, Vincent Vanhoucke, and Alexander~A Alemi.
\newblock Inception-v4, inception-resnet and the impact of residual connections
  on learning.
\newblock In \emph{AAAI}, volume~4, page~12, 2017.

\bibitem[Teerapittayanon et~al.(2016)Teerapittayanon, McDanel, and
  Kung]{teerapittayanon2016branchynet}
Surat Teerapittayanon, Bradley McDanel, and HT~Kung.
\newblock Branchynet: Fast inference via early exiting from deep neural
  networks.
\newblock In \emph{Pattern Recognition (ICPR), 2016 23rd International
  Conference on}, pages 2464--2469. IEEE, 2016.

\bibitem[Viola and Jones(2004)]{viola2004robust}
Paul Viola and Michael~J Jones.
\newblock Robust real-time face detection.
\newblock \emph{{IJCV}}, 57\penalty0 (2):\penalty0 137--154, 2004.

\bibitem[Wang et~al.(2016)Wang, Chang, Yang, Liu, and Huang]{wang2016studying}
Zhangyang Wang, Shiyu Chang, Yingzhen Yang, Ding Liu, and Thomas~S Huang.
\newblock Studying very low resolution recognition using deep networks.
\newblock In \emph{Proceedings of the IEEE Conference on Computer Vision and
  Pattern Recognition}, pages 4792--4800, 2016.

\bibitem[Wei et~al.(2018)Wei, Shen, Cheng, Shi, Xiong, Feng, and
  Huang]{wei2018ts2c}
Yunchao Wei, Zhiqiang Shen, Bowen Cheng, Honghui Shi, Jinjun Xiong, Jiashi
  Feng, and Thomas Huang.
\newblock Ts2c: Tight box mining with surrounding segmentation context for
  weakly supervised object detection.
\newblock In \emph{{ECCV}}, 2018.

\bibitem[Xie et~al.(2018)Xie, Wang, Zhang, Lai, Hong, and
  Qi]{xie2018interleaved}
Guotian Xie, Jingdong Wang, Ting Zhang, Jianhuang Lai, Richang Hong, and
  Guo-Jun Qi.
\newblock Interleaved structured sparse convolutional neural networks.
\newblock In \emph{Proceedings of the IEEE Conference on Computer Vision and
  Pattern Recognition}, pages 8847--8856, 2018.

\bibitem[Xie et~al.(2017)Xie, Girshick, Doll{\'a}r, Tu, and
  He]{xie2017aggregated}
Saining Xie, Ross Girshick, Piotr Doll{\'a}r, Zhuowen Tu, and Kaiming He.
\newblock Aggregated residual transformations for deep neural networks.
\newblock In \emph{{IEEE CVPR}}, pages 5987--5995. IEEE, 2017.

\bibitem[Yang et~al.(2017)Yang, Li, Ouyang, Li, and Wang]{yang2017learning}
Wei Yang, Shuang Li, Wanli Ouyang, Hongsheng Li, and Xiaogang Wang.
\newblock Learning feature pyramids for human pose estimation.
\newblock In \emph{Proceedings of the IEEE International Conference on Computer
  Vision}, pages 1281--1290, 2017.

\bibitem[Yu and Huang(2019)]{yu2019universally}
Jiahui Yu and Thomas Huang.
\newblock Universally slimmable networks and improved training techniques.
\newblock \emph{arXiv preprint arXiv:1903.05134}, 2019.

\bibitem[Yu et~al.(2018)Yu, Yang, Xu, Yang, and Huang]{yu2018slimmable}
Jiahui Yu, Linjie Yang, Ning Xu, Jianchao Yang, and Thomas Huang.
\newblock Slimmable neural networks.
\newblock \emph{arXiv preprint arXiv:1812.08928}, 2018.

\bibitem[Zhang et~al.(2007)Zhang, Zelinsky, and Samaras]{zhang2007real}
Wei Zhang, Gregory Zelinsky, and Dimitris Samaras.
\newblock Real-time accurate object detection using multiple resolutions.
\newblock In \emph{2007 IEEE 11th International Conference on Computer Vision},
  pages 1--8. IEEE, 2007.

\bibitem[Zhang et~al.(2017)Zhang, Zhou, Lin, and Sun]{zhang2017shufflenet}
Xiangyu Zhang, Xinyu Zhou, Mengxiao Lin, and Jian Sun.
\newblock Shufflenet: An extremely efficient convolutional neural network for
  mobile devices.
\newblock \emph{arXiv preprint arXiv:1707.01083}, 2017.

\bibitem[Zhao et~al.(2017)Zhao, Shi, Qi, Wang, and Jia]{zhao2017pyramid}
Hengshuang Zhao, Jianping Shi, Xiaojuan Qi, Xiaogang Wang, and Jiaya Jia.
\newblock Pyramid scene parsing network.
\newblock In \emph{Proceedings of the IEEE conference on computer vision and
  pattern recognition}, pages 2881--2890, 2017.

\bibitem[Zoph et~al.(2018)Zoph, Vasudevan, Shlens, and Le]{zoph2017learning}
Barret Zoph, Vijay Vasudevan, Jonathon Shlens, and Quoc~V Le.
\newblock Learning transferable architectures for scalable image recognition.
\newblock In \emph{{IEEE CVPR}}, 2018.

\end{thebibliography}
\end{document}